\RequirePackage{snapshot}
\documentclass[runningheads]{llncs}

\usepackage{graphicx}
\usepackage{amsmath,amssymb,amsfonts}

\newcommand{\nR}{\mathbb{R}}

\usepackage{hyperref}
\usepackage{csquotes}
\usepackage{caption}
\usepackage{subcaption}

\usepackage{tikz}

\newcommand{\uram}[1]{\marginpar{\tiny\textcolor{blue}{Martin:  #1}}}
\newcommand{\mleh}[1]{\marginpar{\tiny\textcolor{orange}{M.Leh.: #1}}}
\newcommand{\swin}[1]{\marginpar{\tiny\textcolor{magenta}{S.Wint.: #1}}}
\newcommand{\sthu}[1]{\marginpar{\tiny\textcolor{violet}{StHu:  #1}}}
\renewcommand{\uram}[1]{}
\renewcommand{\sthu}[1]{}
\renewcommand{\mleh}[1]{}
\renewcommand{\swin}[1]{}

\usepackage[acronym]{glossaries}
\usepackage{glossaries-extra}

\usepackage{color}
\setabbreviationstyle[acronym]{long-short}
\newacronym{tsp}{TSP}{Traveling Salesperson Problem}
\newacronym{ai}{AI}{Artificial Intelligence}
\newacronym{ga}{GA}{Genetic Algorithm}
\newacronym{hca}{HC}{Hill Climbing}

\usepackage{cleveref}

\begin{document}
\title{%
  Improvements for mlrose applied to the Traveling Salesperson Problem%
  \thanks{This preprint has not undergone peer review or any post-submission
  improvements or corrections. The Version of Record of this contribution is
  published in LLNCS, and is available online at
  \url{https://doi.org/10.1007/978-3-031-25312-6_72}.
      }
}

\titlerunning{Improvements for mlrose applied to the Traveling Salesperson Problem}
\author{
    Stefan Wintersteller\inst{1} \and
    Martin Uray\inst{1,2} \and
    Michael Lehenauer\inst{1} \and
	Stefan Huber\inst{1,2}}
\authorrunning{S. Wintersteller et al.}
\institute{Department for Information Technologies and Digitalisation \and
    Josef Ressel Centre for Intelligent and Secure Industrial Automation\\
    Salzburg University of Applied Sciences, Salzburg, Austria \\
	\email{\{
            swintersteller.its-m2020,
            martin.uray,
            mlehenauer.its-m2020,
            stefan.huber
           \}@fh-salzburg.ac.at}}
\maketitle              %

\section*{Abstract}

In this paper we discuss the application of \gls{ai}
to the
exemplary industrial use case of the two-dimensional commissioning problem in
a high-bay storage, which essentially can be phrased as an instance of
\gls{tsp}.

We investigate the \textit{mlrose} library that provides an \gls{tsp} optimizer based on
various heuristic optimization techniques. Our focus is on two methods, namely
\gls{ga} and \gls{hca}, which are provided by \textit{mlrose}.
We present improvements for both methods that yield shorter tour
lengths, by moderately exploiting the problem structure of \gls{tsp}. That is,
the proposed improvements have a generic character and are not limited to
\gls{tsp} only.

\section*{IEEE keywords}
Artificial Intelligence, Traveling Salesperson Problem, Genetic Algorithm, Hill
Climbing, Commissioning, Material Flow

\section{Introduction}

In this paper, we investigate the application of methods of \gls{ai} to an
industrial problem on the example of optimizing commissioning tasks in a
high-bay storage. Our goal is not to improve on the state of the art in this
task but instead shed light on this problem from an \gls{ai} engineering point
of view. From this point of view, we first have to translate this problem
adequately to apply methods of artificial intelligence, then we would seek for
established software implementations of these methods and evaluate these on the
given task of high-bay storage commissioning.

The commissioning problem or order picking problem is the following: We are
given a high-bay storage where goods are stored in slots arranged on a
two-dimensional wall. An order comprises a finite set of places on that wall
that need to be visited to pick up the goods. We desire to do this as quickly
as possible. Assuming a tapping point where the collection device starts and
ends its job, we can interpret this as an instance of the \gls{tsp}: Given a
set of $n$ locations $p_0, \dots, p_{n-1}$ in the plane, we ask for the
shortest closed tour on $\nR^2$ that visits all points $p_0, \dots, p_{n-1}$.

What we essentially ask for is the optimal order at which we visit the
locations $p_i$. Furthermore, the way we measure distances between pairs of
locations is relevant. To sum up, we consider $p_0, \dots, p_{n-1}$ in a metric
space $(X, d)$ with a metric $d$, encode a tour as a permutation $\pi \colon \{0,
\dots, n-1\} \to \{0, \dots, n-1\}$ and ask for a tour $\pi$ that minimizes the
tour length $\ell(\pi)$ with
\begin{align}
  \ell(\pi) = \sum_{i = 0}^{n-1} d(p_{\pi(i)}, p_{\pi((i+1) \bmod n)}).
\end{align}
In this paper, we may interchangeably represent a permutation $\pi$ as the
sequence $(\pi(0), \dots, \pi(n-1))$, when it fits better to the formal
setting.

A natural choice for $d$ is the Euclidean metric, which we use for experiments
in this paper. However, the discussed methods work with any metric and in
practice certain restrictions in the motion of the highbay storage may be
reflected by a respective choice of $d$, such as the Manhattan metric.

\subsection{Related work}

The \gls*{ga} is comprehensively described in \cite{russel2010}. Several improvements, modifications
and adaption for the vanilla implementation for the problem of \gls{tsp} have been proposed, like
by employing Ant Colonies \cite{ga_with_ant_colony},
 Reinforcement Learning and supervised learning \cite{gambardella_ant-q_1995}, or
 recurrent neural networks \cite{tarkov_solving_2015}.
A recent and comprehensive overview on \gls{tsp} using \gls{ai} is given by Osaba et al.
\cite{osaba_traveling_2020}, covering the \gls{ga}. This work highlights the most notable
\gls{ga} crossover variants.

Similarly, also the \gls*{hca} is comprehensively described in Russel and Norvig, including
several modifications to overcome issues, like plateaus or ridges \cite{russel2010}.
Additional extensions, like Simulated Annealing, Tabu Search, the Greedy Randomize
Adaptive Search Procedure, Variable Neighborhood Search, and the Iterated Local Search
support to overcome the local optima problem \cite{al-betar_beta_2017}.

\section{Experimental Setup}

  For all the experiments in this work a common setup is established.
  As there are already libraries for standard implementations for  the \gls{ga}
  and \gls{hca} algorithm,
 we do not implement the algorithms from scratch, rather we use a library called \textit{mlrose}\footnote{\url{https://mlrose.readthedocs.io/}}  as a base and
 improve the above stated algorithm based on this library.
  This library already provides a mapping of the \gls{tsp} to
 a set of implementations of well-known \gls{ai} methods, which makes it a
 favorable candidate for our commissioning task from an engineering point of
 view. For this work however, only the implementation of the \gls{ga} and
 \gls{hca} are used.

 During experiments, an implementation error %
 was
 discovered, which caused \textit{mlrose} to consistently select unfit
 individuals when fitness can assume negative values, which is the case for
 mlrose's implementation for \gls{tsp}%
 \footnote{\url{https://github.com/gkhayes/mlrose/issues/63}}.
 The experiments presented in this paper in particular contain a comparison of
 the original and the fixed version of \textit{mlrose}.

 Our evaluations are based on the well-known precalculated data set
 \textit{att48} from TSPLIB~\cite{Rein91}. This data set contains 48 cities in
 a coordinate system with a known minimal tour length of $33523$ (unit-less).
 All experiments are evaluated using the CPU clock, and were conducted on
 a Intel Core i7-7700K (4.20 GHz).

\section{\glsfirst{ga}}

\subsection{General basics}

The \gls{ga} is an optimization and search procedure that is inspired by the
maxim \enquote{survival of the fittest} in natural evolution. A candidate
solution (individual) is encoded by a string over some alphabet (genetic code).
Individuals are modified by two genetic operators: (i) random alteration
(mutation) of single individuals and (ii) recombination of two parents
(crossover) to form offsprings. Given a set of individuals (population), a
selection mechanism based on a fitness function together with the two genetic
operators produce a sequence of populations (generations). The genetic
operators promote exploration of the search space while the selection mechanism
attempts to promote the survival of fit individuals over generations. The
\gls{ga} as implemented in \textit{mlrose} terminates after no progress has been made
for a certain number of generations or a predefined maximum number of
generations.

For \gls{ga} to work well, it is paramount that a reasonable genetic
representation of individuals is used. In particular, the crossover operator
needs to have the property that the recombination of two fit parents produces
fit offsprings again, otherwise the genetic structure of fit individuals would
not survive over generations and \gls{ga} easily degenerates to a randomized
search.
For further details on the \gls{ga}, the reader be reffered to \cite[Chapter~4]{russel2010}.

\subsection{Implementation in \textit{mlrose}}

The state vector representation %
is directly used by \textit{mlrose}
as genetic representation, i.e., an individual is encoded as a permutation
sequence $\pi$ of the integers $0, \dots, n-1$, which are indices of the $n$
locations to be visited. Recombination of a first parent $\pi_1$ and a second
parent $\pi_2$ works as follows: The sequence $\pi_1$ is considered to be split
at a random position, the prefix of $\pi_1$ is taken and the missing locations
in the genetic string are taken from $\pi_2$ in the order as they appear in
$\pi_2$.

Note that \gls{tsp} has the symmetry property that a solution candidate $\pi$
and its reverse counterpart $\pi^*$ can be considered to be the same solutions.
Not only is $\ell(\pi) = \ell(\pi^*)$ but in some sense the structure of the
solution is the same. The reason behind this is that the pairwise
distances between locations in the Euclidean plane (or adequate metric spaces)
are invariant with respect to reflection.

However, the recombination strategy does not take this symmetry property into
account.
This leads to the following problem: Consider the recombination of two parents,
$\pi_1$ and $\pi_2$, that are reasonably similar and fit, however, their
direction of traversal is essentially opposite. Then the offspring first
traverses the locations like $\pi_1$ and then continues with $\pi_2$ that
possesses the reversed direction, which likely destroys the fit solution
structure displayed by $\pi_1$ and $\pi_2$. That is, two fit parents produce
unfit offsprings.%

As an illustrative extreme example, assume $\pi_1$ is a globally optimal
solution of \gls{tsp} and $\pi_2 = \pi_1^*$. For sake of argument, assume
$\pi_1 = [0, 1, \dots, 7]$ and $\pi_2 = [7, 6, \dots, 0]$.
Then the offspring $\pi_3$ that results from a split in
the middle of the genetic string would be $\pi_3 = [0, \dots, 3, 7, \dots, 4]$,
which is now typically far from globally optimal, i.e., $\ell(\pi_3) \gg
\ell(\pi_1)$.
This recombination would only not
hurt if the middle of the fit tour $\pi_1$, where the split point of the
recombination is located, would happen to be close to the start or end of
$\pi_1$.

\subsection{Modification}

To mitigate the presented issue of the recombination strategy, we
would like to have a natural notion of direction of traversal of a tour, so we
could figure out whether we would need to reverse the parent $\pi_2$ before
recombining it with $\pi_1$. But since we lack an adequate mathematical notion,
we factor out the two possibilities of tour traversals of $\pi_2$ in a
different way.

When recombining $\pi_1$ and $\pi_2$, we actually consider two candidate
offsprings: offspring $\pi_3$ from $\pi_1$ and $\pi_2$ and offspring $\pi_4$
from $\pi_1$ and $\pi_2^*$.
We then compare the fitness values of the two candidate offsprings,
i.e., we compare $\ell(\pi_3)$ and $\ell(\pi_4)$, and keep only the better one
as the recombination result.  Following our observation from the previous
section, we expect that one offspring of two fit parents results from a
direction-conforming recombination and the other does not. (Of course, it still
can happen that the two parents are bad mates for other reasons, i.e., they can
still be structurally insufficiently compatible.)

This way we turn the original recombination operator into a reversal-invariant
recombination operator. Note that our proposed recombination operator is
beneficial not only for \gls{tsp}, but generally for all problems with this
reversal symmetry of the genetic encoding of individuals.

In literature other recombination methods, based on crossover
\cite{roy_novel_2019,hussain_genetic_2017} and mutation
\cite{abdoun_analyzing_2012}, can be found. These recombination operators work
differently, but can all be applied additionally within the proposed method,
instead of the implemented recombination operator.

\subsection{Results}
An experiment was carried out to measure the performance
of the modified \gls{ga} and the fixed version, in comparison
with the default implementation by \textit{mlrose}. All experiments
algorithms use a shared parameter configuration of a population size of $100$
and a maximum number of $300$ generations, if not stated otherwise.
While we would set a small positive mutation rate in practice, the higher the
mutation rate is, the closer both implementations converge to a random search.
Hence, for the sake of comparison, we set the mutation rate to $0.0$ for our
experiment. Carrying out all experiments $1000$ times, the results visually look like Gaussian distributed (not shown in
the paper) and hence the notation of (mean$\pm$std. deviation) is used in the
following discussion.

\begin{figure}[t]
	\centering
	\begin{subfigure}[b]{0.49\textwidth}
		\centering
		\includegraphics[width=\textwidth]{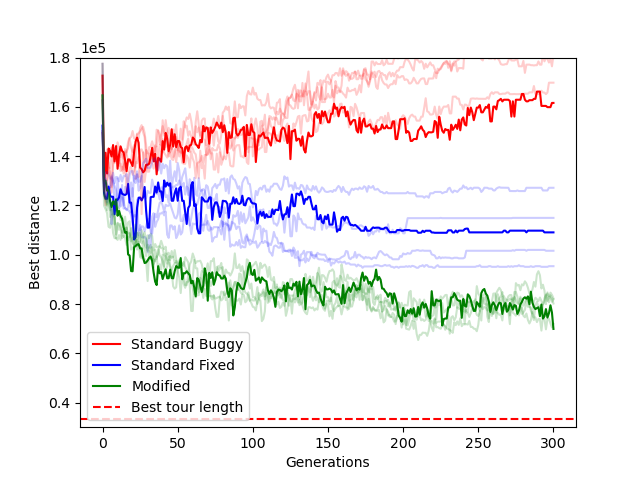}
		\caption{Plots on the tour lengths for the experiments with the
			\gls{ga}. The optimal solution is depicted by the red dashed line. Additional seeds are indicated transparent.}
		\label{fig:gen_distance_plot}
	\end{subfigure}
	\hfill
	\begin{subfigure}[b]{0.49\textwidth}
		\centering
		\includegraphics[width=\textwidth]{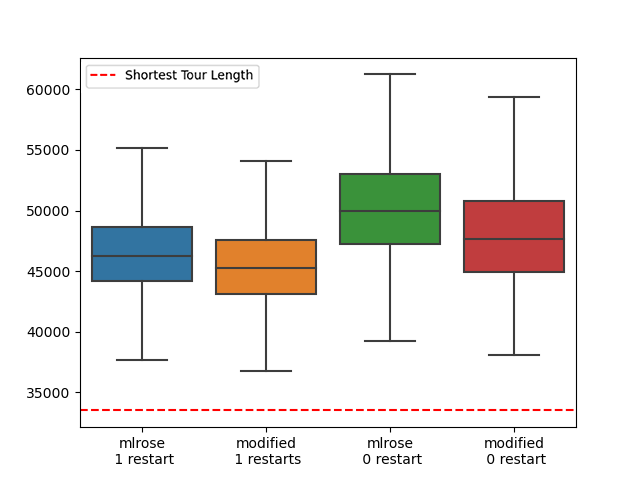}
		\caption{The box plots on the tour lengths for the experiments with the
			\gls{hca} algorithm. The optimal solution is depicted by the red dashed line.}
		\label{fig:hill_distance_plot}
	\end{subfigure}
	\caption{Results of the experiments on the \gls{ga} (\figurename~\ref{fig:gen_distance_plot}) and \gls{hca} (\figurename~\ref{fig:hill_distance_plot}).}
	\label{fig:three graphs}
\end{figure}

The tour lengths of the original implementation (buggy), the fixed,
and modified version are shown over five different initial seed settings
(\cref{fig:gen_distance_plot}). The red dashed horizontal line marks the optimal
solution at a tour length of $33523$.
The results over $1000$ experiments show, that
the modified \gls{ga} ($67961 \pm 3602$) is closer to the optimum than the original algorithm and the fixed version by \textit{mlrose} ($125019 \pm 4438$ and $110137 \pm 6677$, respectively) (however note, that the original version acts againts convergence to a minimum). The modifications pose a decrease in the mean tour lengths by a factor of $0.54$ and $0.61$ for the original and the fixed version, respectively.

The gain in the performance by the modified algorithm is at the expense of a
higher computation time.
The original algorithm by \textit{mlrose} ($18484 \pm 111$ ms) and its fixed version
($18024 \pm 100$ ms) are faster than the modified \gls{ga} ($47988 \pm 161$ ms). The modification of
the \gls{ga} results in a mean computation slowdown by a factor of about $2.59$ (original algorithm) and $2.66$ (fixed algorithm),
compared to the original implementation by \textit{mlrose}.

\section{\glsfirst{hca}}

\subsection{General basics and Implementation in \textit{mlrose}}

\gls{hca} is a simple and most widely known optimization technique,
cf.~\cite{russel2010}. When phrased as a maximization (minimization, resp.)
problem of a function $\ell$ over some domain, Hill Climbing moves stepwise upward
(downward, resp.) along the steepest ascent (descent, resp.) until it reaches a
local maximum (minimum, resp.) of $\ell$.
The domain is
given by the transposition graph $G = (V, E)$ over the vertex set $V$ of
permutations $\pi$, where $E$ contains an edge $(\pi, \pi')$ between vertices
$\pi$ and $\pi'$ iff we can turn $\pi$ into $\pi'$ via a single transposition,
i.e., the tour $\pi'$ results from the tour $\pi$ by swapping two locations
only. %

In more detail, \gls{hca} starts with a random tour $\pi$ and calculates the
cost $\ell(\pi)$ to be minimized. Then it considers all neighbors of $\pi$
within $G$ and their costs.
If the neighbor $\pi'$ with minimum costs has a lower cost than $\ell(\pi)$
then it moves to $\pi'$ and repeats. Otherwise, $\pi$ constitutes a local
minimum and \gls{hca} either terminates or restarts with a new random tour
$\pi$, as implemented in \textit{mlrose}. After a given maximum number of restarts,
\gls{hca} returns the best permutation found in all runs.
For further details on the \gls{hca}, the reader be reffered to \cite[Chapter~4]{russel2010}.

While vanilla \gls{hca} is a simple optimization technique, various
improvements are known in the literature to overcome different shortcomings,
see \cite{russel2010} for an overview.
First of all, \gls{hca} gets stuck in local optima.
To mitigate this issue, the \textit{mlrose} implementation provides a restart mechanism. %
A second well-known issue for \gls{hca} is the existence of plateaus, i.e.,
subregions of the domain where the fitness function is constant such that
\gls{hca} is no uphill direction. Allowing a certain number of sideways
moves \cite{russel2010} would mitigate this issue, while no such mechanism
is implemented in \textit{mlrose}.

\subsection{Modification}

From the discussion in the previous section, we take away that local optima are
the prominent issue for \gls{hca} on \gls{tsp} in \textit{mlrose}.

Here a natural measure for the prominence of a local maximum is the number of
steps in the transposition graph. A prominence of $k$ means that we have to
admit $k$ downward steps from a local maximum until we can pursue an ascending
path that allows us to escape the local maximum.

In the \textit{mlrose} implementation, \gls{hca} is already stuck at local maxima with a
prominence of only $1$, and it would resort to a restart. Our modification
simply allows for a single downward step from local maxima to overcome local
maxima of prominence $1$. If the following step would lead us back to the old local
maximum, we terminate this run and apply a restart as the original version.
More generally, we simply keep a data structure of previously visited
permutations to disallow cycles in the paths traced by \gls{hca}.

This modification leads to another advantage: In the course of restarts, a
series of Hill Climbing searches from randomly generated starting points is
performed. After a restart, if the algorithm reaches a state that was visited
in a previous trial, the Hill Climbing implemented by \textit{mlrose} will take the same
path again and will reach the very same local minimum again. The modified
algorithm terminates after an already visited state is reached, which
constitutes an early out optimization.

\subsection{Results}
An experiment over $1000$ attempts was performed to measure the performance of
the modified \gls{hca} in comparison with the \gls{hca} implementation by
\textit{mlrose}. The tour lengths of the modified algorithm and the \gls{hca}
implemented by \textit{mlrose} using $0$ and $1$ restarts are shown in
\cref{fig:hill_distance_plot}. The results again look visually rather Gaussian
distributed (not shown in the paper), such that we use the notation of
(mean$\pm$standard deviation) in the following. The red dashed horizontal line
again marks the optimal tour length of $33523$.

With $1$ restart, the modified algorithm performs best ($45420 \pm 3340$),
while the implementation by \textit{mlrose} has a slightly higher overall tour length
($46438 \pm 3427$), which is expected since the modified version effectively
extends the exploration.

For further comparison, the restarts parameter of both algorithms (standard
\gls{hca} implementation and modified) are limited to $0$ to test the
performance against the default configuration of \textit{mlrose}. Here the improvement
of the modified version ($47944 \pm 4348$) (red) over the original version
($50263 \pm 4498$) (green) becomes more significant.

The modification influences on the computing time as the modified
algorithm is slightly slower ($36588\pm 4243$ ms) than the \gls{hca}
implemented by \textit{mlrose} ($34128 \pm 3956$ ms). Similar, the results for the
experiments with $0$ restarts. The modified algorithm has a higher computing
time ($18273 \pm 2980$ ms) than the \gls{hca} implementation by \textit{mlrose} ($16701
\pm 2551$ ms).
Increasing the number of restarts from $0$ to $1$ gives a slowdown of a factor
of $1.9$, for the modified and the original implementations likewise.

\section{Conclusion and final remarks}

This paper was motivated by the industrial application of \gls{ai} to the
industrial problem of optimizing commissioning tasks in a high-bay storage,
which translates to the \gls{tsp}.

For the experimental evaluation of the proposed approaches,
we chose \textit{mlrose} for a \gls{ai} library that already provides optimization
routines for \gls{tsp}. With the experiments we had a closer look at two
optimization techniques,
namely \gls{ga} and \gls{hca}. After exploiting and fixing an implementation error within the
library, the problem structure of
\gls{tsp} is analyzed: one improvement for the \gls{ga} and one for the
\gls{hca} are introduced, respectively.
The results show a reduction of $46\%$/$39\%$
for \gls{ga} and $2.1\%$/$4.6\%$ for \gls{hca}. The modifications we propose, however,
have some generic character and are not only applicable to \gls{tsp}.

For the \gls{ga}, a significant improvement on the computed tour length can be
shown based on our reversal-invariant crossover operator. %

For the \gls{hca}, the goal of the experiment was to show, that a problem-specific
treatment is necessary for \gls{tsp}. %
By altering the vanilla implementation towards the properties of the \gls{tsp},
a clear improvement can be observed.

Finally, we would like to remark that to some extent our paper could be seen as
a showcase that \gls{ai} libraries should only carefully be applied as
plug-and-play solutions to industrial problems and the specific problem
structure of the industrial problem at hand likely provides means to improve
the performance of the generic implementations. While the democratization
through meta-learning facilities like AutoML relieve an application engineer
from the tedious search for Machine Learning methods and their hyperparameters for a
given problem at hand, we believe that in general, they do not make an
understanding of the underlying methods obsolete.

\vspace*{-0.125cm}
\subsubsection{Acknowledegments} M. Uray is funded by the Science and Innovation
Strategy Salzburg (WISS 2025) project ``DaSuMa''
(grant number 20204-WISS/140/572/3-2022) and S. Huber by the European
Interreg Österreich-Bayern project AB292 ``KI-Net''.
Both authors are also supported by the Christian Doppler Research Association
(JRC ISIA).
\vspace*{-0.125cm}

\bibliographystyle{splncs04}
\bibliography{bibliography}

\end{document}